\documentclass[11pt,a4paper]{article}
\usepackage[hyperref]{emnlp-ijcnlp-2019}
\usepackage{times}
\usepackage{latexsym}

\usepackage{url}
\usepackage{booktabs}
\usepackage{inputenc}
\usepackage{graphicx}
\usepackage{amsmath}
\usepackage{amssymb}
\usepackage{amsfonts}
\usepackage{multirow}
\usepackage{comment}
\usepackage{subfigure}

\aclfinalcopy 

\title{Multilingual and Multi-Aspect Hate Speech Analysis}

\author{Nedjma Ousidhoum, Zizheng Lin, Hongming Zhang, Yangqiu Song, Dit-Yan Yeung \\
  Department of Computer Science and Engineering \\
  The Hong Kong University of Science and Technology \\    
   \texttt{nousidhoum@cse.ust.hk, zlinai@connect.ust.hk, hzhangal@cse.ust.hk,}\\ \texttt{yqsong@cse.ust.hk, dyyeung@cse.ust.hk} \\
  }

\date{}

\begin{document}
\maketitle
\begin{abstract}

Current research on hate speech analysis is typically oriented towards monolingual and single classification tasks. In this paper, we present a new multilingual multi-aspect hate speech analysis dataset and use it to test the current state-of-the-art multilingual multitask learning approaches. We evaluate our dataset in various classification settings, then we discuss how to leverage our annotations in order to improve hate speech detection and classification in general.

\end{abstract}

\section{Introduction}

With the expanding amount of text data generated on different social media platforms, current filters are insufficient to prevent the spread of hate speech. Most internet users involved in a study conducted by the Pew Research Center report having been subjected to offensive name calling online or witnessed someone being physically threatened or harassed online.\footnote{http://www.pewinternet.org/2017/07/11/online-harassment-2017/} Additionally, Amnesty International within Element AI have lately reported that many women politicians and journalists are assaulted every 30 seconds on Twitter.\footnote{https://www.amnesty.org.uk/press-releases/women-abused-twitter-every-30-seconds-new-study} This is despite the Twitter policy condemning the promotion of violence against people on the basis of race, ethnicity, national origin, sexual orientation, gender identity, religious affiliation, age, disability, or serious disease.\footnote{https://help.twitter.com/en/rules-and-policies/hateful-conduct-policy} 
Hate speech may not represent the general opinion, yet it promotes the dehumanization of people who are typically from minority groups ~\cite{soral2017,Martin2013} and can incite hate crime \cite{RossRCCKW17}.

Moreover, although people of various linguistic backgrounds are exposed to hate speech~\cite{W17-3012,RossRCCKW17}, English is still at the center of existing work on toxic language analysis. 
Recently, some research studies have been conducted on languages such as German~\cite{Kra2017b}, Arabic~\cite{DBLP:conf/asunam/AlbadiKM18}, and Italian~\cite{DBLP:conf/lrec/SanguinettiPBPS18}. 
However, such studies usually use monolingual corpora and do not contrast, or examine the correlations between online hate speech in different languages. On the other hand, tasks involving more than one language such as the hatEval task\footnote{https://competitions.codalab.org/competitions/19935}, which covers English and Spanish, include only separate classification tasks, namely (a) women and immigrants as target groups, (b) individual or generic hate and, (c) aggressive or non-aggressive hate speech. 

\begin{figure*}[t]
\centering
    \subfigure[English.]{\includegraphics[width=0.30\textwidth]{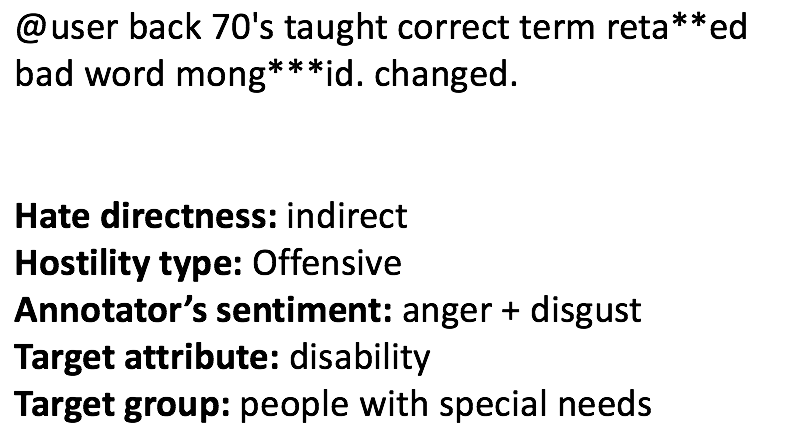} \label{fig:en-annotate-example}
    }
    \subfigure[French.]{\includegraphics[width=0.33\textwidth]{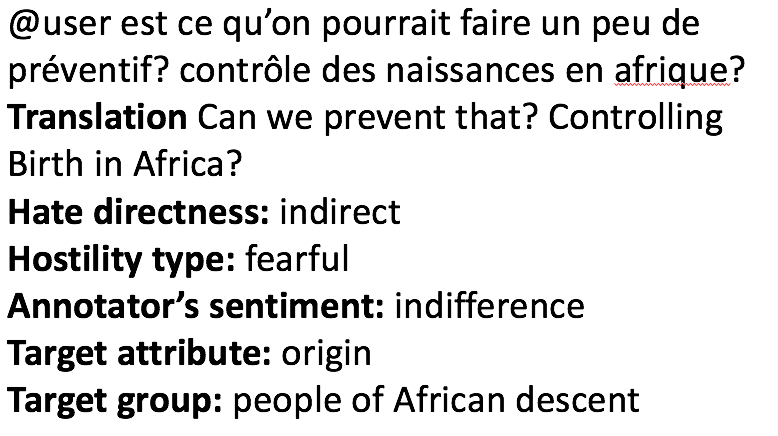} \label{fig:fr-annotate-example}
    }
    \subfigure[Arabic.]{\includegraphics[width=0.33\textwidth]{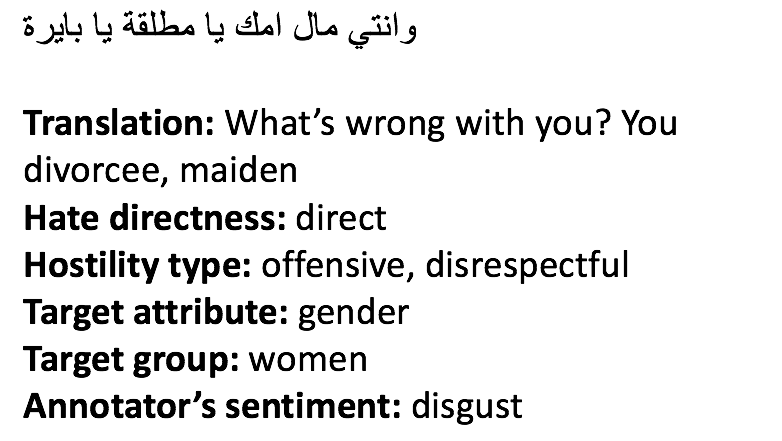} \label{fig:ar-annotate-example}
    }
\vspace{-0.1in}
\caption{Annotation examples in our dataset.}
\vspace{-0.1in}
\label{fig:annotation-example}
\end{figure*}

Treating hate speech classification as a binary task may not be enough to inspect the motivation and the behavior of the users promoting it and, how people would react to it.
For instance, the hateful tweets presented in Figure~\ref{fig:annotation-example} show toxicity directed towards different targets, with or without using slurs, and generating several types of reactions.
We believe that, in order to balance between truth and subjectivity, there are at least five important aspects in hate speech analysis. Hence, our annotations indicate (a) whether the text is direct or indirect; (b) if it is offensive, disrespectful, hateful, fearful out of ignorance, abusive, or normal; (c) the attribute based on which it discriminates against an individual or a group of people; (d) the name of this group; and (e) how the annotators feel about its content within a range of negative to neutral sentiments.
To the best of our knowledge there are no other hate speech datasets that attempt to capture fear out of ignorance in hateful tweets or examine how people react to hate speech. We claim that our multi-aspect annotation schema would provide a valuable insight into several linguistic and cultural differences and bias in hate speech.

We use Amazon Mechanical Turk to label around 13,000 potentially derogatory tweets in English, French, and Arabic based on the above mentioned aspects and, regard each aspect as a prediction task.
Since in natural language processing, there is a peculiar interest in multitask learning, where different tasks can be used to help each other~\cite{DBLP:journals/jmlr/CollobertWBKKK11,ruder2017sluice,Joint_Multitask_D17-1206}, we use a unified model to handle the annotated data in all three languages and five tasks. We adopt~\cite{ruder2017sluice} as a learning algorithm adapted to loosely related tasks such as our five annotated aspects and, use the Babylon cross-lingual embeddings~\cite{babylon_Smith17} to align the three languages.
We compare the multilingual multitask learning settings with monolingual multitask, multilingual single-task, and monolingual single-task learning settings respectively. Then, we report the performance results of the different settings and discuss how each task affects the remaining ones.
We release our dataset and code to the community to extend research work on multilingual hate speech detection and classification.\footnote{our code is available on: \url{https://github.com/HKUST-KnowComp/MLMA_hate_speech} } 



\begin{table*}[t]
\centering
 \small
\begin{tabular}{l|c|c|r}
\toprule
Dataset & \# Tweets & Labels & Annotators/Tweet \\ \midrule
\citet{Chatzakou:2017:MBD:3091478.3091487} & 9,484 &aggressive, bullying, spam, normal & 5 \\\hline
\citet{DBLP:conf/naacl/WaseemH16} & 16, 914 & racist, sexist, normal & 1 \\ \hline
\citet{DavidsonWMW17} & 24, 802 & hateful, offensive (but not hateful), neither & 3 or more \\ \hline
\multirow{2}{*}{\citet{Golbeck2017}} & \multirow{2}{*}{35,000}  & the worst, threats, hate speech, direct & \multirow{2}{*}{2 to 3} \\ & & harassment, potentially offensive, non-harassment &  \\ \hline
\multirow{2}{*}{\citet{FountaDCLBSVSK18}} & \multirow{2}{*}{80, 000} & offensive, abusive, hateful speech, & \multirow{2}{*}{5 to 20} \\ & & aggressive, cyberbullying, spam, normal &  \\ \hline
\multirow{2}{*}{\citet{hatelingo}} & \multirow{2}{*}{28,608} & directed, generalized + target = archaic, class, disability, &  \multirow{2}{*}{3} \\ & &  ethnicity, gender, nationality, religion, sexual orientation &  \\ \midrule 
Ours & 13,000 & Labels for five different aspects & 5
\\ \bottomrule 
\end{tabular}
\caption{Comparative table of some of the available hate speech and abusive language corpora in terms of labels and sizes.} 
\label{comparative_tablel}
\vspace{-0.1in}
\end{table*}

\section{Related Work}

\begin{figure}[!t]
     \centering
     \includegraphics[scale=0.4]{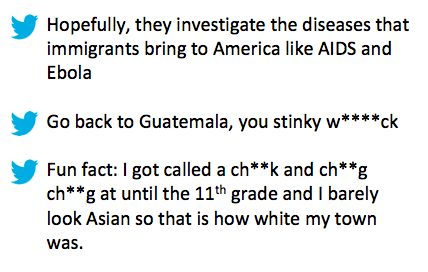}
     \caption{Three tweets in which (1) the first one accuses immigrants of harming society without using any direct insult; (2) the second insults a Hispanic person using a slur; and (3) the third one uses slurs to give a personal account. This shows that profanity is not a clear indicator of the presence of hate speech.}
     \label{fig:en_hate_speech_example}
 \end{figure}

There is little consensus on the difference between profanity and hate speech and, how to define the latter \cite{W17-1101}. As shown in Figure~\ref{fig:en_hate_speech_example}, slurs are not an unequivocal indicator of hate speech and can be part of a non-aggressive conversation, while some of the most offensive comments may come in the form of subtle metaphors or sarcasm~\cite{MalmasiZ18}. Consequently, there is no existing human annotated vocabulary that explicitly reveals the presence of hate speech, which makes the available hate speech corpora sparse and noisy \cite{Nobata:2016:ALD:2872427.2883062}. 

Given the subjectivity and the complexity of such data, annotation schemes have rarely been made fine-grained. Table~\ref{comparative_tablel} compares different labelsets that exist in the literature. For instance, ~\citet{DBLP:conf/naacl/WaseemH16} use racist, sexist, and normal as labels; ~\citet{DavidsonWMW17} label their data as hateful, offensive (but not hateful), and neither, while~\citet{hatelingo} present an English dataset that records the target category based on which hate speech discriminates against people, such as ethnicity, gender, or sexual orientation and ask human annotators to classify the tweets as hate and non hate. \citet{FountaDCLBSVSK18} label their data as offensive, abusive, hateful, aggressive, cyberbullying, spam, and normal. On the other hand, \citet{D18-1391} have chosen to detect ideologies of hate speech counting 40 different hate ideologies among 13 extremist hate groups.

The detection of hate speech targets is yet another challenging aspect of the annotation. \citet{park2018_genderbias} report the bias that exists in the current datasets towards identity words, such as \textit{women}, which may later cause false predictions. They propose to debias gender identity word embeddings with additional data for training and tuning their binary classifier. We address this false positive bias problem and the common ambiguity of target detection by asking the annotators to label target attributes such as origin, gender, or religious affiliation within 16 named target groups such as refugees, or immigrants.
 
Furthermore,~\citet{KLUBICKA18.2} have reproduced the experiment of~\citet{DBLP:conf/naacl/WaseemH16} in order to study how hate speech affects the popularity of a tweet, but discovered that some tweets have been deleted. For replication purposes, we provide the community with anonymized\footnote{In conformity with Twitter terms and conditions.} tweet texts rather than IDs.
 
Non-English hate speech datasets include Italian, German, Dutch, and Arabic corpora.~\citet{DBLP:conf/lrec/SanguinettiPBPS18}~present a dataset of Italian tweets, in which the annotations capture the degree of intensity of offensive and aggressive tweets, in addition to whether the tweets are ironic and contain stereotypes or not. \citet{RossRCCKW17}~have collected more than 500 German tweets against refugees, and annotated them as hateful and not hateful. \citet{DBLP:conf/ranlp/HeeLVMDPDH15}~detect bullies and victims among youngsters in Dutch comments on AskFM, and classify cyberbullying comments as insults or threats. Moreover, \citet{DBLP:conf/asunam/AlbadiKM18}~provide a corpus of Arabic sectarian speech. 

Another predominant phenomenon in hate speech corpora is code switching. \citet{hindi_english_hatespeech2018} present a dataset of code mixed Hindi-English tweets, while \citet{galery2018}~report the presence of Hindi tokens in English data and use multilingual word embeddings to deal with this issue when detecting toxicity. Similarly, we use such embeddings to take advantage of the multilinguality and comparability of our corpora during the classification. 

Our dataset is the first trilingual dataset comprising English, French, and Arabic tweets that encompasses various targets and hostility types. Additionally, to the best of our knowledge, this is the first work that examines how annotators react to hate speech comments.
 
To fully exploit the collected annotations, we tested multitask learning on our dataset. Multitask learning~\cite{DBLP:journals/jmlr/CollobertWBKKK11} allows neural networks to share parameters with one another and, thus, learn from related tasks. It has been used in different NLP tasks such as parsing \cite{Joint_Multitask_D17-1206}, dependency parsing \cite{peng2017mtl_dependency}, neural machine translation \cite{mtl_mt2016}, sentiment analysis \cite{augenstein2018mtl}, and other tasks. Multitask learning architectures tackle challenges that include sharing the label space and the question of private and shared space for loosely related tasks \cite{ruder2017sluice}, for which techniques may involve a massive space of potential parameter sharing architectures.

\section{Dataset}
In this section, we present our data collection methodology and annotation process.
\subsection{Data Collection}
Considering the cultural differences and commonly debated topics in the main geographic regions where English, French, and Arabic are spoken, searching for equivalent terms in the three languages led to different results at first. Therefore, after looking for 1,000 tweets per 15 more or less equivalent phrases in the three languages, we revised our search words three times by questioning the results, adding phrases, and taking off unlikely ones in each of the languages.
In fact, we started our data collection by searching for common slurs and demeaning expressions such as \textit{``go back to where you come from''}. Then, we observed that discussions about controversial topics, such as \textit{feminism} in general, \textit{illegal immigrants} in English, \textit{Islamo-gauchisme} (``Islamic leftism") in French, or \textit{Iran} in Arabic were more likely to provoke disputes, comments filled with toxicity and thus, notable insult patterns that we looked for in subsequent search rounds. 
\subsection{Linguistic Challenges}

All of the annotated tweets include original tweets only, whose content has been processed by (1) deleting unarguably detectable spam tweets, (2) removing unreadable characters and emojis, and (3)  masking the names of mentioned users using \textit{@user} and potentially enclosed URLs using \textit{@url}. As a result, annotators had to face the lack of context generated by this normalization process.

Furthermore, we perceived code-switching in English where Hindi, Spanish, and French tokens appear in the tweets. Some French tweets also contain Romanized dialectal Arabic tokens generated by, most likely, bilingual North African Twitter users. Hence, although we eliminated most of these tweets in order to avoid misleading the annotators, the possibly remaining ones still added noise to the data. 

One more challenge that the annotators and ourselves had to tackle, consisted of Arabic diglossia and switching between different Arabic dialects and Modern Standard Arabic (MSA). While MSA represents the standardized and literary variety of Arabic, there are several Arabic dialects spoken in North Africa and the Middle East in use on Twitter. Therefore, we searched for derogatory terms adapted to different circumstances, and acquired an Arabic corpus that combines tweets written in MSA and Arabic dialects. For instance, the tweet shown in Figure~\ref{fig:annotation-example} contains a dialectal slur that means ``maiden.''

\subsection{Annotation Process}
We rely on the general public opinion and common linguistic knowledge to assess how people view and react to hate speech.\footnote{We have also provided the annotators with the Urban Dictionary definitions of some slang English words they may not be aware of.} Given the subjectivity and difficulty of the task, we reminded the annotators not to let their personal opinions about the topics being discussed in the tweets influence their annotation decisions.

Our annotation guidelines explained the fact that offensive comments and hate do not necessarily come in the form of profanity. Since different degrees of discrimination work on the dehumanization of individuals or groups of people in distinct ways, we chose not to annotate the tweets within two or three classes. For instance, a sexist comment can be disrespectful, hateful, or offensive towards women. Our initial labelset was established in conformity with the prevalent anti-social behaviors people tend to deal with. We also chose to address the problem of false positives caused by the misleading use of identity words by asking the annotators to label both the target attributes and groups. 

\paragraph{Avoiding scams}
To prevent scams, we also prepared three annotation guideline forms and three aligned labelsets written in English, French, and Modern Standard Arabic with respect to the language of the tweets to be annotated.

We requested native speakers to annotate the data and chose annotators with good reputation scores (more than 0.90). We informed the annotator in the guidelines, that in case of noticeable patterns of random labeling on a substantial number of tweets, their work will be rejected and we may have to block them. Since the rejection affects the reputation of the annotators and their chances to get new tasks on Amazon Mechanical Turk, well-reputed annotators are usually reliable. We have divided our corpora into smaller batches on Amazon Mechanical Turk in order to facilitate the analysis of the annotations of the workers and, fairly identify any incoherence patterns possibly caused by the use of an automatic translation system on the tweets, or the repetition of the same annotation schema. If we reject the work of a scam, we notify them, then reassign the tasks to other annotators.

\subsection{Pilot Dataset}
We initially put samples of 100 tweets in each of the three languages on Amazon Mechanical Turk. We showed the annotators the tweet along with lists of labels describing (a) whether it is direct or indirect hate speech; (b) if the tweet is dangerous, offensive, hateful, disrespectful, confident or supported by some URL, fearful out of ignorance, or other; (c) the target attribute based on which it discriminates against people, specifically, race, ethnicity, nationality, gender, gender identity, sexual orientation, religious affiliation, disability, and other (``other'' could refer to political ideologies or social classes.); (d) the name of its target group, and (e) whether the annotators feel anger, sadness, fear or nothing about the tweets. 

Each tweet has been labeled by three annotators. We have provided them with additional text fields to fill in with labels or adjectives that would (1) better describe the tweet, (2) describe how they feel about it more accurately, and (3) name the group of people the tweet shows bias against. We kept the most commonly used labels from our initial labelset, took off some of the initial class names and added frequently introduced labels, especially the emotions of the annotators when reading the tweets and the names of the target groups. For instance, after this step, we have ended up merging \textit{race, ethnicity, nationality} into one label \textit{origin} given common confusions we noticed and; added \textit{disgust} and \textit{shock} to the emotion labelset; and introduced \textit{socialists} as a target group label since many annotators have suggested these labels.

\subsection{Final Dataset}
The final dataset is composed of a pilot corpus of 100 tweets per language, and comparable corpora of 5,647 English tweets, 4,014 French tweets, and 3,353 Arabic tweets. Each of the annotated aspects represents a classification task of its own, that could either be evaluated independently, or, as intended in this paper, tested on how it impacts other tasks. The different labels are designed to facilitate the study of the correlations between the explicitness of the tweet, the type of hostility it conveys, its target attribute, the group it dehumanizes, how different people react to it, and the performance of multitask learning on the five tasks.
We assigned each tweet to five annotators, then applied majority voting to each of the labeling tasks. Given the  numbers of annotators and labels in each annotation sub-task, we allowed multilabel annotations in the most subjective classification tasks, namely the hostility type and the annotator's sentiment labels, in order to keep the right human-like approximations. If there are two annotators agreeing on two labels respectively, we add both labels to the annotation.

The average Krippendorff scores for inter-annotator agreement (IAA) are 0.153, 0.244, and 0.202 for English, French, and Arabic respectively, which are comparable to existing complex annotations~\cite{DBLP:conf/lrec/SanguinettiPBPS18} given the nature of the labeling tasks and the number of labels.

We present the labelset the annotators refer to, and statistics of our annotated data in the following.

\begin{table}[t!]
    \centering
{\small
\begin{tabular}{l|l|l|l|l}
\toprule
Attribute & Label & En & Fr & Ar \\ \midrule
\multirow{2}{*}{Directness}            & Direct             & 530     & 2,198   & 1,684   \\
                                       & Indirect           & 4,456    & 997   & 754   \\ \midrule
\multirow{6}{*}{Hostility}                 & Abusive            & 671	  & 1,056  & 610     \\
                                       & Hateful            & 1,278	  & 399       & 755       \\
                                       & Offensive          & 4,020		        & 1,690       &  1,151      \\
                                       & Disrespectful      & 782    		      & 396       & 615        \\
                                       & Fearful            & 562        & 388      &  41      \\
                                       & Normal         & 1,359	     & 1,124       & 1,197       \\ \midrule
\multirow{6}{*}{Target}                & Origin             &  2,448 & 2,266  	& 877     \\
                                       & Gender             & 638	& 27	& 548  \\
                                       & Sexual Orientation &   514		      & 12       & 0       \\
                                       & Religion           &    	68		     & 146       & 145       \\
                                       & Disability         &  1,089		       & 177       & 1      \\
                                       & Other              &  890		      & 1,386       &  1,782      \\ \midrule
\multirow{5}{*}{Group} & Individual            &  497	& 918	& 915       \\
                                       & Other              &   1,590 &	1,085 &	1,470 \\
                                       & Women              &  878	& 62	& 722     \\
                                       & Special needs            &  1,571	& 174	& 2   \\
                                       & African descent      &  86	& 311	& 51 \\ \midrule
\multirow{7}{*}{Annotator} & Disgust            &  3,469       & 602  & 778       \\
                                       & Shock              &   	2,151	& 1,179	& 917 \\
                                       & Anger              &  2,955       & 531	       & 356       \\
                                       & Sadness            &  2,775	& 1,457	& 388   \\
                                       & Fear               &  1,304       & 378	& 35 \\
                                       & Confusion          & 1,747        &  446      & 115       \\ 
                                       & Indifference       & 2,878        & 2,035        & 1,825       \\ \midrule
\multicolumn{2}{l|}{Total number of tweets}                 & 5,647	& 4,014	& 3,353 \\ \bottomrule
\end{tabular}
}
\caption{The label distributions of each task. The counts of direct and indirect hate speech include all tweets except those that are single labeled as ``normal". Tweet and annotator's sentiment (Annotator) are multilabel classification tasks, while target attribute (Target) and target group (Group) are not.}
    \label{tab:statistics_table}
    \vspace{-0.1in}
\end{table}

\paragraph{Directness label}

Annotators determine the explicitness of the tweet by labeling it as \textit{direct} or \textit{indirect} speech. This should be based on whether the target is explicitly named, or less easily discernible, especially if the tweet contains humor, metaphor, or figurative speech. Table~\ref{tab:statistics_table} shows that even when partly using equivalent keywords to search for candidate tweets, there are still significant differences in the resulting data. 

\paragraph{Hostility type}
To identify the hostility type of the tweet, we stick to the following conventions: (1) if the tweet sounds dangerous, it should be labeled as \textit{abusive}; (2) according to the degree to which it spreads hate and the tone its author uses, it can be \textit{hateful}, \textit{offensive} or \textit{disrespectful}; (3) if the tweet expresses or spreads fear out of ignorance against a group of individuals, it should be labeled as \textit{fearful}; (4) otherwise it should be annotated as \textit{normal}. We define this task to be multilabel. Table~\ref{tab:statistics_table} shows that hostility types are relatively consistent across different languages and offensive is the most frequent label.

\paragraph{Target attribute}
After annotating the pilot dataset, we noticed common misconceptions regarding race, ethnicity, and nationality, therefore we merged these attributes into one label \textit{origin}. Then, we asked the annotators to determine whether the tweet insults or discriminates against people based on their (1) \textit{origin}, (2) \textit{religious affiliation}, (3) \textit{gender}, (4) \textit{sexual orientation}, (5) \textit{special needs} or (6) \textit{other}. Table~\ref{tab:statistics_table} shows there are fewer tweets targeting disability in Arabic compared to English and French and no tweets insulting people based on their sexual orientation which may be due to the fact that the labels of gender, gender identity, and sexual orientation use almost the same wording. On the other hand, French contains a small number of tweets targeting people based on their \textit{gender} in comparison to English and Arabic. We have observed significant differences in terms of target attributes in the three languages. More data may help us examine the problems affecting targets of different linguistic backgrounds.

\paragraph{Target group}
We determined 16 common target groups tagged by the annotators after the first annotation step. The annotators had to decide on whether the tweet is aimed at \textit{women, people of African descent, Hispanics, gay people, Asians, Arabs, immigrants in general, refugees}; people of different religious affiliations such as  \textit{Hindu, Christian, Jewish people, and Muslims}; or from political ideologies \textit{socialists, and others}. We also provided the annotators with a category to cover hate directed towards one \textit{individual}, which cannot be generalized.
In case the tweet targets more than one group of people, the annotators should choose the group which would be the most affected by it according to them.
Table~\ref{comparative_tablel} shows the counts of the five categories out of 16 that commonly occur in the three languages. In fact, most of the tweets target individuals or fall into the ``other'' category. In the latter case, they may target people with different political views such as liberals or conservatives in English and French, or specific ethnic groups such as Kurdish people in Arabic. 
English tweets tend to have more tweets targeting people with special needs, due to common language-specific demeaning terms used in conversations where people insult one another. Arabic tweets contain more hateful comments towards women for the same reason. On the other hand, the French corpus contains more tweets that are offensive towards African people, due to hateful comments generated by debates about immigrants.

\paragraph{Sentiment of the annotator}
We claim that the choice of a suitable emotion representation model is key to this sub-task, given the subjective nature and social ground of the annotator's sentiment analysis. After collecting the annotation results of the pilot dataset regarding how people feel about the tweets, and observing the added categories, we adopted a range of sentiments that are in the negative and neutral scales of the hourglass of emotions introduced by~\citet{CambriaHourglassEmotions}. This model includes sentiments that are connected to objectively assessed natural language opinions, and excludes what is known as self-conscious or moral emotions such as shame and guilt. Our labels include \textit{shock}, \textit{sadness}, \textit{disgust}, \textit{anger}, \textit{fear}, \textit{confusion} in case of ambivalence, and \textit{indifference}. This is the second multilabel task of our model.

Table~\ref{tab:statistics_table} shows more tweets making the annotators feel disgusted and angry in English, while annotators show more indifference in both French and Arabic. A relatively more frequent label in both French and Arabic is shock, therefore reflecting what some of the annotators were feeling during the labeling process.

\section{Experiments}

We report and discuss the results of five classification tasks: (1) the directness of the speech, (2) the hostility type of the tweet, (3) the discriminating target attribute, (4) the target group, and (5) the annotator's sentiment.

\subsection{Models}

We compare both traditional baselines using bag-of-words (BOW) as features on Logistic regression (LR), and deep learning based methods.

For deep learning based models, we run bidirectional LSTM (biLSTM) models with one hidden layer on each of the classification tasks. Deeper BiLSTM models performed poorly due to the size of the tweets. 
We chose to use Sluice networks~\cite{ruder2017sluice} since they are suitable for loosely related tasks such as the annotated aspects of our corpora.

We test different models, namely single task single language (STSL), single task multilingual (STML), and multitask multilingual models (MTML) on our dataset.
In multilingual settings, we tested Babylon multilingual word embeddings~\cite{babylon_Smith17} and MUSE~\cite{lample2017unsupervised} on the different tasks. We use Babylon embeddings since they appear to outperform MUSE on our data.

Sluice networks \cite{ruder2017sluice} learn the weights
of the neural networks sharing parameters (sluices) jointly with the rest of the model and share an embedding layer, Babylon embeddings in our case, that associates the elements of an input sequence.
We use a standard 1-layer BiLSTM partitioned into two subspaces, a shared subspace and a private one, forced to be orthogonal through a regularization penalty term in the loss function in order to enable the multitask network to learn both task-specific and shared representations.
The hidden layer has a dimension of 200, the learning rate is initially set to 0.1 with a learning rate decay, and we use the DyNet \cite{dynet} automatic minibatch function to speed-up the computation. 
We initialize the cross-stitch unit to imbalanced, set the standard deviation of the Gaussian noise to 2, and use simple stochastic gradient descent (SGD) as the optimizer. 

All compared methods use the same split as train:dev:test=8:1:1 and the reported results are based on the test set.
We use the dev set to tune the threshold for each binary classification problem in the multilabel classification settings of each task.

\begin{table*}[t!]
\centering
{\small
\begin{tabular}{l|l|l|l|l|l|l|l|l|l}
\toprule
\multirow{2}{*}{Attribute}    & \multirow{2}{*}{Model} & \multicolumn{4}{c}{Macro-F1} & \multicolumn{4}{|l}{Micro-F1} \\  \cline{3-10} 
          &  & EN & FR & AR & Avg & EN & FR & AR & Avg    \\ \midrule
\multirow{7}{*}{Directness}  
		  & Majority & 0.50 & 0.11 & 0.50 & 0.47 & 0.79 & 0.41 & 0.54 & 0.58   \\
          & LR & 0.52 & 0.50 & 0.53 & 0.52 & 0.79 & 0.50 & 0.56 & 0.62          \\
          & STSL & \textbf{0.94} & \textbf{0.80} & \textbf{0.84} & \textbf{0.86} & \textbf{0.89} & \textbf{0.69} & \textbf{0.72} & \textbf{0.76} \\
          & MTSL & \textbf{0.94} & 0.65 & 0.76 & 0.78 & \textbf{0.89} & 0.58 & 0.65 & 0.70          \\
          & STML & \textbf{0.94} & 0.79 & 0.83 & 0.85 & 0.88 & 0.66 & \textbf{0.72} & 0.75          \\
          & MTML & \textbf{0.94} & 0.78 & 0.74 & 0.82 & 0.88 & 0.66 & 0.65 & 0.73          \\ \bottomrule
          \end{tabular}
          }
            \caption{Full evaluation scores of the only binary classification task where the single task single language model consistently outperforms multilingual multitask models. }
              \label{tab:directness_evaluation_table}
    \vspace{-0.1in}
\end{table*}

\begin{table*}[t!]
    \centering
{\small

\begin{tabular}{l|l|l|l|l|l|l|l|l|l}
\toprule
\multirow{2}{*}{Attribute}    & \multirow{2}{*}{Model} & \multicolumn{4}{c}{Macro-F1} & \multicolumn{4}{|l}{Micro-F1} \\  \cline{3-10} 
          &  & EN & FR & AR & Avg & EN & FR & AR & Avg    \\ \midrule

\multirow{7}{*}{Tweet}   
		  & Majority & 0.24 & 0.19 & 0.20 & 0.21 & 0.41 & 0.27 & 0.27 & 0.32   \\
          & LR & 0.14 & 0.20 & 0.25 & 0.20 & 0.54 & 0.56 & \textbf{0.48} & 0.53          \\
          & STSL & 0.24 & 0.12 & 0.31 & 0.23 & 0.49 & 0.51 & 0.47 & 0.49  \\
		  & MTSL & 0.09 & 0.20 & 0.33 & 0.21 & \textbf{0.55} & \textbf{0.59} & 0.46 & \textbf{0.54}  \\
		  & STML & 0.04 & 0.07 & \textbf{0.35} & 0.16 & 0.54 & 0.47 & 0.37 & 0.46  \\
		  & MTML &\textbf{0.30} &\textbf{0.28} &\textbf{0.35} &\textbf{0.31} &0.45 &0.48 &0.44 &0.46 \\ \midrule
\multirow{7}{*}{Target Attribute}   
		  & Majority & 0.15 & 0.13 & 0.28 & 0.19 & 0.25 & 0.32 & 0.40 & 0.32   \\
		  & LR & 0.41 & 0.35 & 0.47 & 0.41 & 0.52 & 0.55 & 0.53 & 0.53          \\
          & STSL & 0.42 & 0.18 & \textbf{0.63} & 0.41 & \textbf{0.68} & 0.71 & 0.50 & 0.63          \\
          & MTSL & 0.41 & \textbf{0.43} & 0.41 & \textbf{0.42} & \textbf{0.68} & 0.67 & \textbf{0.56} & \textbf{0.64} \\
          & STML & 0.39 & 0.09 & 0.24 & 0.24 & 0.67 & 0.62 & 0.53 & 0.61          \\
          & MTML & \textbf{0.43} & 0.24 & 0.16 & 0.28 & 0.66 & \textbf{0.72} & 0.51 & 0.63         \\ \midrule
\multirow{7}{*}{Target Group} 
		  & Majority & 0.07 & 0.06 & 0.08 & 0.07 & 0.18 & 0.14 & 0.35 & 0.22   \\
		  & LR & \textbf{0.18} & 0.33 & \textbf{0.40} & \textbf{0.30} & 0.34 & 0.40 & 0.62 & 0.46          \\
          & STSL & 0.04 & 0.21 & 0.04 & 0.10 & 0.48 & \textbf{0.59} & 0.58 & 0.55          \\
          & MTSL & 0.04 & 0.27 & 0.15 & 0.15 & \textbf{0.50} & 0.54 & 0.55 & 0.53          \\
          & STML & 0.11 & \textbf{0.37} & 0.13 & 0.20 & 0.49 & 0.57 & \textbf{0.64} & \textbf{0.56} \\
          & MTML & 0.06 & 0.19 & 0.10 & 0.11 & \textbf{0.50} & 0.54 & 0.56 & 0.53              \\ \midrule
\multirow{7}{*}{Annotator's Sentiment}  
		  & Majority & 0.42 & 0.21 & 0.17 & 0.27 & 0.46 & 0.31 & 0.32 & 0.39   \\
		  & LR   & 0.29 & 0.15 & 0.14 & 0.19 & 0.45 & 0.30 & 0.46 & 0.40          \\
		  & STSL & \textbf{0.57} & \textbf{0.30} & 0.12 & \textbf{0.33} & 0.57 & 0.39 & \textbf{0.48} & 0.48 \\
		  & MTSL & \textbf{0.57} & 0.17 & 0.17 & 0.30 & 0.57 & \textbf{0.50} & 0.45 & 0.51 \\
		  & STML & 0.47 & 0.22 & 0.13 & 0.27 & \textbf{0.59} & 0.49 & \textbf{0.48} & \textbf{0.52} \\
		  & MTML & 0.55 & 0.20 & \textbf{0.21} & 0.32 & 0.58 & 0.45 & 0.45 & 0.49 \\ 
		  \bottomrule
\end{tabular}
}
    \caption{Full evaluation of tasks where multilingual and multitask models outperform on average single task single language model on four different tasks.}
    \label{tab:evaluation_table}
    \vspace{-0.1in}
\end{table*}

\subsection{Results and Analysis}
We report both the micro and macro-F1 scores of the different classification tasks in Tables \ref{tab:directness_evaluation_table} and \ref{tab:evaluation_table}. \textit{Majority} refers to labeling based on the majority label, \textit{LR} to logistic regression, \textit{STSL} to single task single language models, \textit{STML} to single task multilingual models, and \textit{MTML} to multitask multilingual models.
\paragraph{STSL}
STSL performs the best among all models on the directness classification, and it is also consistent in both micro and macro-F1 scores. 
This is due to the fact that the directness has only two labels and multilabeling is not allowed in this task. 
Tasks involving imbalanced data, multiclass and multilabel annotations harm the performance of the directness in multitask settings.

Since macro-F1 is the average of all F1 scores of individual labels, all deep learning models have high macro-F1 scores in English which indicates that they are particularly good at classifying the \textit{direct} class.
STSL is also comparable or better than traditional BOW feature-based classifiers when performed on other tasks in terms of micro-F1 and for most of the macro-F1 scores. This shows the power of the deep learning approach.
\paragraph{MTSL}
Except for the directness, MTSL usually outperforms STSL or is comparable to it. When we jointly train each task on the three languages, the performance decreases in most cases, other than the target group classification tasks. This may be due to the difference in label distributions across languages. 
Yet, multilingual training of the target group classification task improves in all languages. Since the target group classification task involves 16 labels, the amount of data annotated for each label is lower than in other tasks.
Hence, when aggregating annotated data in different languages, the size of the training data also increases, due to the relative regularity of identification words of different groups in all three languages in comparison to other tasks.
\paragraph{MTML}
MTML settings do not lead to a big improvement which may be due to the class imbalance, multilabel tasks, and the difference in the nature of the tasks. In order to inspect which tasks hurt or help one another, we trained multilingual models for pairwise tasks such as (group, target), (hostility, annotator's sentiment), (hostility, target), (hostility, group), (annotator's sentiment, target) and (annotator's sentiment, group). We noticed that when trained jointly, the target attribute slightly improves the performance of the tweet's hostility type classification by 0.03,0.05 and 0.01 better than the best reported scores in English, French, and Arabic, respectively. When target groups and attributes are trained jointly, the macro F-score of the target group classification in Arabic improves by 0.25 and when we train the tweet's hostility type within the annotator's sentiment, we improve the macro F-score of Arabic by 0.02. We believe that we can take advantage of the correlations between target attributes and groups along with other tasks, to set logic rules and develop better multilingual and multitask settings.

\section{Conclusion}
In this paper, we presented a multilingual hate speech dataset of English, French, and Arabic tweets.
We analyzed in details the difficulties related to the collection and annotation of this dataset. 
We performed multilingual and multitask learning on our corpora and showed that deep learning models perform better than traditional BOW-based models in most of the multilabel classification tasks. Multilingual multitask learning also helped tasks where each label had less annotated data associated with it. 

Better tuned deep learning settings in our multilingual and multitask models would be expected to outperform the existing state-of-the-art embeddings and algorithms applied to our data.
The different annotation labels and comparable corpora would help us perform transfer learning and investigate how multimodal information on the tweets, additional unlabeled data, label transformation, and label information sharing may boost the classification performance in the future.

\section*{Acknowledgement}
This paper was supported by the Early Career Scheme (ECS, No. 26206717) from Research Grants Council in Hong Kong, and by postgraduate studentships from the Computer Science and Engineering department of the Hong Kong University of Science and Technology.
\bibliography{emnlp2019.bib}
\bibliographystyle{acl_natbib}
\end{document}